\documentclass[letterpaper, 10 pt, conference]{ieeeconf}  % Comment this line out if you need a4paper
\pdfoutput=1

\usepackage[left=54pt,top=50pt,right=54pt,bottom=57pt]{geometry}

\IEEEoverridecommandlockouts                              % This command is only needed if 
\usepackage[normalem]{ulem}

\usepackage[caption=false]{subfig}

\usepackage[utf8]{inputenc}
\usepackage{moreverb,url}
\usepackage{times}
\usepackage{multicol}
\usepackage{amsmath,amsfonts,amssymb,mathtools,leftidx}%,amsthm}
\usepackage{wrapfig}
\usepackage{float}
\usepackage{xcolor}
\usepackage{graphicx}
\usepackage{multirow}
\usepackage{mathrsfs}
\usepackage{nccmath}
\usepackage{balance}
\usepackage{afterpage}

\usepackage{tabularx}
\usepackage[noadjust]{cite}
\usepackage[colorlinks,bookmarksopen,bookmarksnumbered,citecolor=red,urlcolor=red]{hyperref}
\usepackage{booktabs}
\usepackage[ruled,vlined,linesnumbered,algo2e]{algorithm2e} % for algorithm

\usepackage{algorithm,algpseudocode}
\usepackage[a-1b]{pdfx}
\usepackage[colorinlistoftodos,prependcaption,textsize=tiny]{todonotes}

\makeatletter
\newcommand\fs@spaceruled{\def\@fs@cfont{\bfseries}\let\@fs@capt\floatc@ruled
  \def\@fs@pre{\vspace{8pt}\par\noindent\rule[2pt]{\columnwidth}{1pt}}%height.8pt depth0pt
  \def\@fs@post{\vspace{1pt}\par\noindent\rule[0pt]{\columnwidth}{1pt}}%
  \def\@fs@mid{\vspace{0pt}\par\noindent\rule[2pt]{\columnwidth}{1pt}}%
  \let\@fs@iftopcapt\iftrue}
\makeatother

\title{\LARGE \bf
BabyNet: A Lightweight Network for Infant Reaching\\ Action Recognition in Unconstrained Environments to\\ Support Future Pediatric Rehabilitation Applications
}
\author{Amel Dechemi,$^1$ Vikarn Bhakri,$^1$ Ipsita Sahin,$^2$ Arjun Modi,$^2$ Julya Mestas,$^2$ Pamodya Peiris,$^1$ \\ Dannya Enriquez Barrundia,$^2$ Elena Kokkoni,$^2$ and Konstantinos Karydis$^1$
\thanks{$^1$ Dept. of Electrical and Computer Engineering, University of California, Riverside. 
	Email: {\{adech003, vbhak001, ppeir002, karydis\}@ucr.edu}.} 
\thanks{$^2$ Dept. of Bioengineering, University of California, Riverside. 
	Email: {\{isahi001, amodi003, jmest001, denri013, elena.kokkoni\}@ucr.edu}.}
}

\begin{document}

\maketitle
\thispagestyle{empty}
\pagestyle{empty}

%%%%%%%%%%%%%%%%%%%%%%%%%%%%%%%%%%%%%%%%%%%%%%%%%%%%%%%%%%%%%%%%%%%%%%%%%%%%%%%%
\begin{abstract}
Action recognition is an important component to improve autonomy of physical rehabilitation devices, such as wearable robotic exoskeletons. Existing human action recognition algorithms focus on adult applications rather than pediatric ones. In this paper, we introduce BabyNet, a light-weight (in terms of trainable parameters) network structure to recognize infant reaching action from off-body stationary cameras. We develop an annotated dataset that includes diverse reaches performed while in a sitting posture by different infants in unconstrained environments (e.g., in home settings, etc.). Our approach uses the spatial and temporal connection of annotated bounding boxes to interpret onset and offset of reaching, and to detect a complete reaching action. We evaluate the efficiency of our proposed approach and compare its performance against other learning-based network structures in terms of capability of capturing temporal inter-dependencies and accuracy of detection of reaching onset and offset.  Results indicate our BabyNet can attain solid performance in terms of (average) testing accuracy that exceeds that of other larger networks, and can hence serve as a light-weight data-driven framework for video-based infant reaching action recognition.
\end{abstract}

%%%%%%%%%%%%%%%%%%%%%%%%%%%%%%%%%%%%%%%%%%%%%%%%%%%%%%%%%%%%%%%%%%%%%%%%%%%%%%%%
%\vspace{-3pt}
\section{Introduction}
Identifying human motor actions that emerge early in life (e.g., spontaneous movements of arms and legs, kicking, crawling, etc.) is an emerging vision-based action recognition research direction~\cite{Chambers2019ComputerVT, Emeli2020TowardsStudy, 9412822}.  %a significant body of work on action recognition targeted to adult populations~\cite{Moltisanti2019ActionRF,Farha2019MSTCNMT,Crasto2019MARSMR,Si2019AnAE,Stroud2020D3DD3,Sudhakaran2020GateShiftNF,Materzynska2020SomethingElseCA,Ji2020ActionGA}.
The ability to automate the process of detecting, recognizing, and classifying actions performed by young children and infants from visual data can prove useful across applications. 
Examples include monitoring for safety~\cite{Suzuki2012ActivityRF,Goto2013ActivityRS,Nie2018ACC}, studying infants' interaction with caregivers \cite{Klein2020IncorporatingInteraction}, identifying markers for diagnosis of neuromotor disorders \cite{Chambers2019ComputerVT, Emeli2020TowardsStudy, Westeyn2011MonitoringCD, Stahl2012AnOF, Hashemi2012ComputerVT}, as well as--primarily of interest herein--closing action-perception loops for smart environments and assistive wearable robotic devices~\cite{Efthymiou2018MultiVF,Tsiami2018Multi3MP,Kokkoni2020GEARingSE,Marinoiu20183DHS,Kokkoni2020DevelopmentReachingb}. %, as well as autonomous wearable robots for pediatric rehabilitation~\cite{Kokkoni2020DevelopmentReachingb}

Developing algorithms to accurately and reliably recognize infant motor actions is not straightforward. 
There is inherent movement variability within and across young humans as a natural result of learning and growth   \cite{Fetters2010PerspectiveAction.}.
%Infants' bodies have distinct proportions and can attain distinct poses compared to those of adults \cite{Sciortino2017OnTE}.\adc{add a transition on the impact of reaching in infants' development and why the focus}
For example, the kinematic properties of infant reaching (the motor action of interest in this paper) change over the first few years as infants learn how to adapt to environmental, task, and biomechanical constraints \cite{Konczak1997TheLife, Thelen1993TheDynamics.}.
In fact, it may take years for children to achieve smooth and straight reaching trajectories similar to those seen in adults~\cite{Berthier2005DevelopmentOR,Schneiberg2002TheDO}. 
Thus, existing skeletal models or learning-based pose estimation methods often used in adult action recognition may not fit well in pediatric action recognition~\cite{Efthymiou2018MultiVF,Sciortino2017OnTE,Suzuki2012ActivityRF}.
%This was confirmed in previous work, in which models based on adult motor action recognition under-performed when tested with datasets from children \cite{Efthymiou2018MultiVF,Suzuki2012ActivityRF}.

%Another key limitation is the absence of human activity datasets that contain sufficient, detailed, and variable infant body poses. 
State-of-the-art adult action recognition methods require very large training datasets~\cite{Soomro2012UCF101AD,Carreira2017QuoVA,Kuehne2011HMDBAL,Shahroudy2016NTURA,Damen2018ScalingEV,Goyal2017TheS}. %Gupta2020QuoVS,
Because of protection of privacy of minors,  and given that data from adult activities can be used in a wider range of applications compared to infant activities, there is a lack of infant activity datasets. A few notable exceptions exist~\cite{Hesse2018ComputerVF,Hesse2018LearningAI,9412822}; however, these datasets do not apply in this work that specifically targets infant reaching. 
We focus on reaching as it is one of the earliest motor milestones leading to environment exploration and learning, and hence, advancing reaching is an important goal in pediatric rehabilitation~\cite{lobo2013onset, lobo2015characterization}.
In this context, vision-based reaching action recognition may guide autonomous reasoning regarding the amount of passive/active feedback that assistive upper extremity devices may provide to the pediatric user~\cite{Kokkoni2020DevelopmentReachingb}.

This paper has a twofold aim. First, to develop a new 
%To this end, we develop herein a new 
dataset focusing on infant reaching. Second, to use the dataset to develop a machine learning algorithm for infant reaching action recognition. 
The dataset is constructed based on diverse online-shared videos that demonstrate reaching actions of both typically-developing infants as well as infants with arm mobility challenges (all between 6--12 months of age). 
Annotations and bounding boxes that describe reaching properties (e.g., reaching onset/offset, object touched, arm position in beginning/end of reach, etc.) are also inlcuded.
Next, \emph{BabyNet}, a new network structure aimed at infant reaching action recognition is developed.
The network is built upon a long short-term memory (LSTM) module to model different stages of reaching action through a spatial-temporal interpretation. Our motivation is to provide a light-weight (based on the number of trainable parameters) structure of comparable efficiency with (significantly) larger ones.

Succinctly, the contributions of this paper include:
\vspace{-1pt}
\begin{itemize}
    \item A new dataset for infant reaching action recognition including action annotations and bounding boxes. %(To the best of our knowledge, no such dataset is available)
    \vspace{0pt}
    \item A light-weight network \emph{BabyNet} able to model short-range and long-range motion dependencies of the different stages of infant reaching action.
    \vspace{0pt}
    \item Performance validation and comparison of \emph{BabyNet} against other learning-based methods using our dataset.
\end{itemize}

% --------------------------------- %
\section{Related Works}
% --------------------------------- %

\subsection{Overview of Human Action Recognition Approaches}
State-of-the-art video-based human action recognition algorithms have been steadily shifting from the use of Support Vector Machines or Hidden Markov models (e.g.,~\cite{Schldt2004RecognizingHA,Gu2010ActionAG}) to the use of deep learning networks, primarily due to the accessibility, adaptability, accuracy and decrease in time execution the latter can offer (e.g.,~\cite{Simonyan2014TwoStreamCN,Wang2016TemporalSN,Feichtenhofer2016ConvolutionalTN,Donahue2015LongtermRC}).
%
%One of the first significant attempts was from Simonyan and Zisserman~\cite{Simonyan2014TwoStreamCN} who used two separated CNNs (streams) trained to extract features from a sampled RGB video frame paired with the surrounding stack of optical flow images. The prediction scores of both streams are fused using averaging and training a multi-class linear SVM~\cite{Crammer2001OnTA} on stacked L2-normalized softmax scores as features. Other works~\cite{Wang2015TowardsGP,Wang2016TemporalSN,Feichtenhofer2016ConvolutionalTN,Chron2015PCNNPC,Donahue2015LongtermRC,Gkioxari2015FindingAT,Ng2015BeyondSS} followed upon those steps. %For example, Feichtenhofer et al.~\cite{Feichtenhofer2016ConvolutionalTN} built upon the two-stream architecture~\cite{Simonyan2014TwoStreamCN} to develop an architecture able to fuse spatial and temporal cues at several levels of granularity in feature abstraction, and with spatial as well as temporal integration.  
One of the first significant attempts~\cite{Simonyan2014TwoStreamCN} used two separated convolutional neural networks (CNNs) trained to extract features from a sampled RGB video frame paired with the surrounding stack of optical flow images. As optical flow is computationally expensive and has a burdensome optimization process, most follow-on works used a CNN to learn the optical flow prediction~\cite{Hui2018LiteFlowNetAL,Fan2018EndtoEndLO}, thus reducing the number of parameters as only one network is needed. Other methods explored the advantages of LSTM structures~\cite{Donahue2015LongtermRC} to incorporate motion by updating the pooling of the features across time~\cite{Geest2018ModelingTS, Ge2019AnAM}.

%Such approaches use still images/frames extracted from the video as the input, and as result it may be hard to capture motion actions that are established by some longer-horizon temporal correlation. %~\cite{REF?}. 
%
%These approaches take one frame as input and could miss the motion of an action through time. 
%Therefore, 3D convolution networks were proposed to learn spatio-temporal features~\cite{Tran2015LearningSF,Diba2018Temporal3C} by using temporally-densely-sampled sequences of images as the input. One main downside of the 3D structure is a high number of parameters being involved in the model. This results in an increase of computational cost and necessitates large-scale training datasets. In an effort to come up with more light-weight while efficient solutions, some works use (2+1)D decomposition instead of the 3D structure \cite{Diba2018Temporal3C,Xie2018RethinkingSF,7410879,9008828,8953824, He2019StNetLA,9157646} or combine 2D CNN and 3D CNN \cite{9008827,8578773,Xie2018RethinkingSF,ECO_eccv18,9009790, 8578152}

Despite some remarkable results achieved by RGB-video-based methods to date, some key challenges remain. These include, for example, background clutter, illumination disparity, pose/viewpoint variation, to name a few. One way to improve recognition performance under these challenges is via skeleton data representations that do not contain color information. %and thus it is not impacted by the challenges RGB videos are facing. 
Early relevant works did not use information regarding internal dependencies between body joints% and dismissed parts of the information related to target action
~\cite{6909476, Liu2017EnhancedSV,7299176, 7298714}; more recent works apply graph convolutional networks to extract features by building a skeleton graph composed of vertices and edges to represent joints and links, respectively~\cite{Thakkar2018PartbasedGC, 8428616,Yan2018SpatialTG}. 
%
% How the datasets can't be applied infant action recognition 
%\textcolor{red}{Already stated point about dataset, wouldn't be repetitive}
These approaches rely on datasets that contain mostly motion actions performed by adults~\cite{Soomro2012UCF101AD,Carreira2017QuoVA,Kuehne2011HMDBAL,Shahroudy2016NTURA,Damen2018ScalingEV,Goyal2017TheS}. 

\subsection{Action Recognition for Rehabilitation Purposes}

Action recognition algorithms can contribute to the field of rehabilitation through their integration in the automation of assistive devices and for assessment of training outcomes.
A recent paradigm shift seeks to develop technology and perform training in the real world to increase dosage and help translate better the effects of training%since the training will be taking place in the same context as the application 
~\cite{galloway2019innovative,Reinkensmeyer2012TechnologiesDisability}. 
Infants, in particular, learn from exposure to complex (unconstrained) and variable environments, as well as in the presence of immediate rewards and constant motivation as they perform motor tasks in such environments. %\adc{extend the challenges and difficulties? (perhaps in Introduction?)}
%Consequently, training and assessment approaches should be applied in such environmental contexts~\cite{Lee2018TheWalk}.
However, taking training outside of the lab or clinic poses various challenges, such as timely and accurate outcome evaluation, and in many cases, absence of a rehabilitation expert. %thus, there is a need for developing approaches in this context. 
Related works tend to use virtual reality or camera-based systems~\cite{8217765,7034540,6623199,5116559} in support of training in such settings. %Examples of technology applications offering a wide range of training interaction activities through ``serious games" and giving direct access to an objective performance feedback utilize virtual reality or camera systems \cite{8217765,7034540,10.1145/1753326.1753649,6623199,5116559,6721804}. 
%\sout{An example is one of the games which seeks to promote gross arm movements in order to aid with upper limb rehabilitation~\cite{5116559}. To track arm motion, the user either wears a glove or holds on to an object of a single color.}

%Other applications focus more on the monitoring of patients. For instance, Collins et al. proposed a daily activity observation system for stroke patients~\cite{8217765} using depth and skeleton data obtained from a Kinect v2 depth camera to assess motion actions. 
%Along the same lines, the authors in~\cite{6721804} determined the capability of Support Vector Machines (SVM) and Random Forests (RF) to accurately classify Kinect-based kinematic activities in physical rehabilitation.

%Although some work in action recognition for adult rehabilitation exists, there is a lack of similar approaches aimed at pediatric (and especially infant) rehabilitation applications~\cite{Pulido2017EvaluatingTC}.
%Approaches developed for and tested with adults may not be the most appropriate for application to infant action recognition for a number of reasons.
%\kkr{These include differences related to body properties and kinematic patterns.
%Our proposed action recognition method is in line with this approach through the analysis of infant reaching actions in non-sterile and unconstrained environments.% (more information to follow).

Specifically in the context of pediatric rehabilitation, recent paradigms involve action recognition methods to make interaction with assistive technology more efficient and effective~\cite{Kokkoni2020GEARingSE, Efthymiou2018MultiVF, 9412822,Tsiami2018Multi3MP, Pulido2017EvaluatingTC}.
For example, social humanoid robots assist in non-contact upper-limb rehabilitation autonomously, by performing a set of prescribed arm-poses for a child to imitate~\cite{Pulido2017EvaluatingTC}. 
Movements here are captured by a Kinect depth camera, stored as 3D skeletons and then compared with entries from a knowledge base. 
Other applications use multiple cameras to alleviate issues such as occlusions that are more likely to happen in sessions involving young children \cite{Kokkoni2020GEARingSE, Efthymiou2018MultiVF, 9412822,Tsiami2018Multi3MP}.
For example, Kokkoni et al. \cite{Kokkoni2020GEARingSE} developed a learning environment for infants using socially-assistive robots and body weight support technology. 
Kinect cameras were utilized to capture movement, and action recognition algorithms were developed to close the loop between the infants and the robots \cite{Kokkoni2020GEARingSE, 9412822}.
Our proposed action recognition work is in line with the aforementioned approaches but for use with a different type of technology; our future goal is to use reaching action recognition to automate our wearable upper extremity device for infants so as to provide passive/active feedback during reaching~\cite{Kokkoni2020DevelopmentReachingb}.
%Nevertheless, the number and type of infant motor actions that have been utilized in action recognition research still remains limited.\adc{add statements on the wearables?}

%{However, related works on action recognition for infants involve different datasets for evaluation, corresponding to different environments and conditions. Availability of a dataset that can serve as a common basis for infant action recognition is still missing. One of the key contributions of this present paper is to create such as dataset.}

% --------------------------------- %
\section{Methods}
% --------------------------------- %

\subsection{Dataset of Infant Reaching Actions}
%In this work, a new dataset was developed containing a collection of videos and various descriptors of infant reaching. The latter were obtained from the videos through annotation and digitization analyses. 

\begin{figure*}[!ht]
\vspace{6pt}
    \centering
    \includegraphics[scale = 0.8]{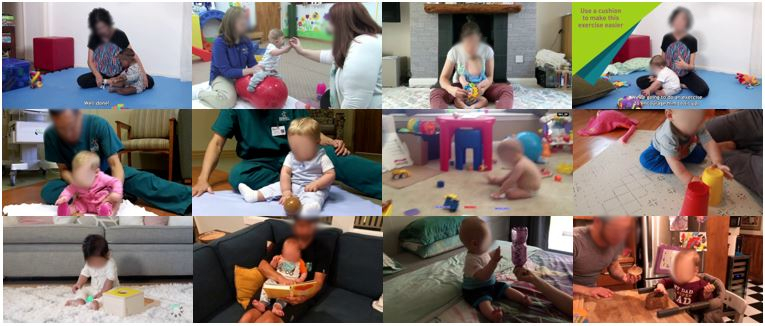}
    \vspace{-6pt}
    \caption{Examples of frames from the dataset.}
    \label{fig:selection_dataset}
    \vspace{-12pt}
\end{figure*}

%\subsubsection{Identification of Videos}
Videos were collected from the YouTube\textsuperscript{TM} online video-sharing platform using search terms such as `infant,' `reaching,' `grabbing,' and `sitting.' 
Videos were included in the dataset if they displayed awake and behaving infants: (1) up to 12 months of age, (2) placed in a sitting position, (3) reaching for objects (regardless of shape and/or position), (4) performing at least one reaching action during which the camera remained stationary, and (5) performing at least one reaching action during which both hand and object were fully visible.
Both typically-developing infants and infants with arm mobility challenges were considered. 
The majority of videos were recorded in natural (unconstrained) environments (e.g., family's home, clinic).
The clothes of infants, presented objects, and the background varied from video to video. Figure~\ref{fig:selection_dataset} shows some illustrative samples.  %snapshots of infants across the different environments.

A total of 193 reaches performed by 21 distinct subjects were collected through 20 videos (Table \ref{tab:Annotation table}).\,\footnote{Video handling procedures were according to YouTube's statement on fair use of videos for research purposes.} %\kkb{  T12/T13/T14 same subject but at different age can we keep them as a different subject?}
In most cases, the video description was detailing the age and gender of the subject; in a few exceptions that this information was omitted (entries marked with an * in Table \ref{tab:Annotation table}), our research team empirically estimated these characteristics. 
%\kkm{Infants' age was verified by one of the co-authors and infants' diagnosis was based on information provided in the video.}\todo{revise}
Out of the 21 subjects, five subjects had a medical diagnosis (four with Down syndrome and one with congenital anomaly; information was provided in the video description). 
Each subject was assigned a unique identification number. 
%that were downloaded using ? software
%What resolution did the videos have?
%total #of reaches??? Are we including the histogram?
%Why didn't we concentrate on infants 4-6 mo only? This is when reaching develops..Maybe say that we kept it more open to ensure that we would get novel and experienced reachers and thus have a richer dataset which also increases variability. Also, it would be hard to identify the exact 4-6 mo age range if not reported in the video description.
%also why did we include both TD and DD infants?
%
%\subsubsection{Video Annotations}\label{reach-annotation}
%\sout{Each video was calibrated by selecting a known measure that was present in every video to adjust for differences in image resolution and camera placement (e.g., angle, distance from infant, etc.). The infant’s cornea was considered to be that point of reference. A horizontal line, drawn from one end of the cornea to the other, was calibrated at 1.05 cm which is the average iris diameter reported for this age range \cite{Ronneburger2006GrowthOT}.}

\begin{table}[!h]
\vspace{6pt}
    \caption{Subjects and Annotated Reaches/Objects per Subject.}
    \vspace{-10pt}
    \begin{center}
    {\footnotesize
    \begin{tabular}{p{1cm}>{\centering}p{1cm}>{\centering}p{1cm}>{\centering}p{0.5cm}>{\centering}p{0.5cm}>{\centering}p{0.5cm}>{\centering\arraybackslash}p{1cm}}
    %\begin{tabular}{c c c c c c c c c c c c c}
    \toprule
        \multirow{1}{*}{Subject} & Age & Gender & \multicolumn{3}{c}{\# Reaches} & \multirow{1}{*}{Total} \\
    \multirow{1}{*}{ID} & [months] & [M / F] & LH & RH & Total & \multirow{1}{*}{Obj.} \\
     \midrule
     \midrule
     T01 & 6-8 & F & 5 & 3 & 8 & 4\\
     \midrule
     T02 & 8-10 & M & 8 & 4 & 12 & 4\\
     \midrule
     T03 & 11-12$^{\ast}$ &M & 2 & 8 & 10 & 2\\
     \midrule
     T04 & 6-12 & F & 3 & 5 & 8 & 5\\ 
     \midrule
     T05 & 10-12 & M$^{\ast}$ & 3 & 1 & 4 & 2\\ 
     \midrule
     T06 & 10-12 & M$^{\ast}$ & 0 & 2 & 2 & 2\\ 
     \midrule
     T07 & 6-7$^{\ast}$ & M& 2 & 3 & 5 & 3\\ 
     \midrule
     T08 & $<$12$^{\ast}$ & M& 4 & 5 & 9 & 1\\
     \midrule
     T09 & 6 & F & 3 & 6 & 9 & 2\\
     \midrule
     T10 & 6 & F & 3 & 3 & 6 & 3\\
     \midrule
     T11 & 6-8 & M& 2 & 1 & 3 & 2\\
     \midrule
     T12 & 8 & F & 16 & 33 & 49 & 8\\
     \midrule
     T13 & 10 & F & 5 & 9 & 14 & 5\\
     \midrule
     T14 & 6 & F & 1 & 1 & 2 & 2\\
     \midrule
     T15 & 9 & M & 20 & 20 & 40 & 10\\
     \midrule
     T16 & 7 & F & 2 & 1 & 3 & 2\\
     \midrule
     \midrule
     D01 & 6-9 & F & 1 & - & 1 & 1\\ 
     \midrule
     D02 & 10 & M & - & 1 & 1 & 1\\
     \midrule
     D03 & $<$12$^{\ast}$ & M & 4 & 1 & 5 & 1\\
     \midrule
     D04 & 9-12 & M & 1 & - & 1 & 1\\ 
     \midrule
     D05 & $<$12$^{\ast}$ & M & 1 & - & 1 & 1\\
     \midrule
     \midrule
    {\bf Total} & -- & -- & {\bf 86} & {\bf 107} & {\bf 193} & {\bf 57}\\
\bottomrule
    \end{tabular}
    }
    \end{center}
    \label{tab:Annotation table}
    \vspace{-20pt}
\end{table}

\begin{figure}[!h]
\vspace{6pt}
    \centering
    %\subfloat[]
    {
    \includegraphics[trim={3cm 0 3cm 0 },clip,scale = 0.30]{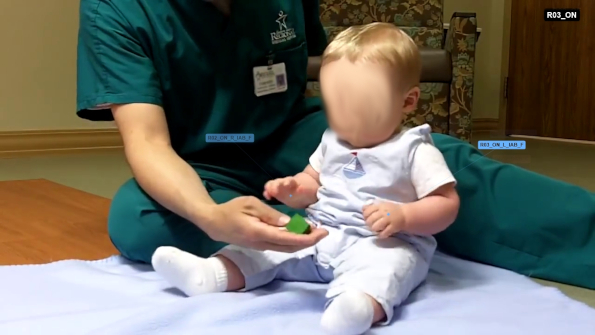}
    \hspace{3pt}
    \includegraphics[trim={3cm 0 3cm 0 },clip,scale = 0.30]{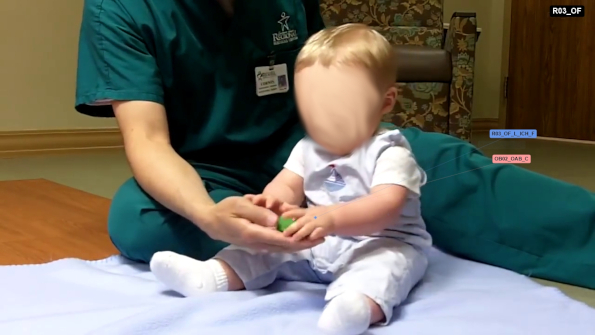}
    }
    %\subfloat[Reaching Offset (RF)]{
    %\includegraphics[trim={3cm 0 3cm 0 },clip,scale = 0.36]{offset-new2.jpg}
    %}
    %\subfloat[]{
    %\includegraphics[trim={3.1cm 0 3.1cm 0 },clip,scale = 0.33]{Acronym2.jpg}
    %
    %\hspace{3pt}
    %\includegraphics[trim={3cm 0 3cm 0 },clip,scale = 0.36]{offset-new2.jpg}
    %}
    \vspace{-6pt}
    \caption{Sample annotated Reaching Onset (left) and Offset (right) frames. %Bottom: Initial hand position and object position was annotated with respect to five body levels depicted here (left). Example of reaching trajectory obtained during digitization analysis (right).
    }
    \label{fig:example_annotation}
    \vspace{-15pt}
\end{figure}

Videos were processed by an open-source software (www.kinovea.org) to obtain annotations of the Reaching Onset (RN) and the Reaching Offset (RF) of every reaching action in all videos. %were annotated. 
RN was defined as the first frame in which the infant’s hand (left, right, or both) started moving toward the presented object. 
RF was defined as the first frame in which the infant's hand intentionally touched the object (Figure~\ref{fig:example_annotation}). 
Hand-to-hand object transfers, partial (interrupted or unfinished) reaching actions, and actions containing frames where the hand and/or object were occluded, were excluded from further analyses in this paper.

Action annotation analyses %(LH/RH reaches and objects) 
were conducted by four researchers of our team. These four researchers first went through a separate protocol to establish reliability.
Reaching actions from two sample videos were initially annotated by all four researchers. 
Only when they all achieved a 100$\%$ agreement on the frequency of reaching actions and a +/-3 frame selection difference between RN and RF, they proceeded with analysis of the remaining videos (two researchers assigned per remaining video). %\kkr{objects annotation not mentioned here} \kkb{  most disagreement was not about the object to be reached but when the reach begins and ends and if it is reach}

%%%%%%%%%%%%%%%%%%%%%%%%%%%%%%%
\begin{figure}[!h]
\vspace{-3pt}
    \centering
    \includegraphics[trim={1cm 0cm 1cm 0.725cm},clip,scale = 0.54]{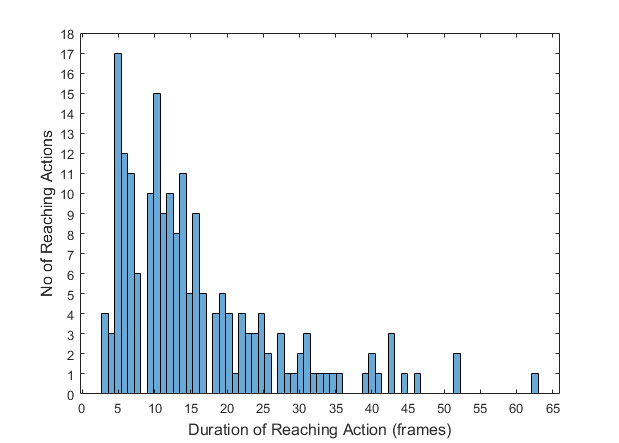}
    \vspace{-8pt}
    \caption{Distribution of reaches duration frequencies.}
    \label{fig:histogram2}
    \vspace{-15pt}
\end{figure}

The annotation process resulted in a total of 193 reaches with 86 reaches performed by the left hand (LH) and 107 by the right hand (RH) and 57 different objects involved in the reaching action (see Table~\ref{tab:Annotation table}). The stage of motor development each subject is at affects both the quality of reaching actions as well as their length. Figure~\ref{fig:histogram2} depicts the distribution of the duration of all reaches in our dataset. The large variability of reaching action duration can be readily observed, though the majority last between 5-15 frames. 

%\kkr{Refer to histogram here?} \kkb{better to mention it after listing the annotation results and then refer to table I and the histogram}

%Table \ref{tab:Annotation table} provides an overview of the actions annotations  %\sout{and the bounding boxes} 
%per subject.

%%%%%%%%%%%%%%%%%%%%%%%%%
%%%%%%%%%%%%%%%%%%%%%%%%%%%%%%%%%%%%%%

\subsection{Annotation of Bounding Boxes}
To assess the spatial and temporal connection during the reaches, we employed detection of each of the hands (LH, RH) and any object(s) on each frame using bounding boxes. 
Two researchers of the team annotated 607 images randomly sampled out of total of 2,984 frames for the complete dataset. For each frame, we created bounding boxes of the infant, their left hand, their right hand, and the objects involved in the reaching action. Annotation of the selected 607 frames resulted in a total of 3,194 bounding boxes. %that were annotated. 

Using the sampled images and corresponding bounding boxes we trained an object detector to help automate the process for the remaining frames. We started with a pre-trained YOLOv3 object detector \cite{Redmon2018YOLOv3AI} with the COCO dataset \cite{Lin2014MicrosoftCC}, which we then finetuned with our obtained annotations. During finetuning, four categories were registered for the detector: infant, left hand (LH), right hand (RH) and object. The object detector was trained and tested using a 75\% (training) / 25\% (testing) split.

\subsection{Baseline Approaches for Comparison}
We considered four baseline network structures to serve as the basis to evaluate our proposed structure's performance. 
%
%The four baseline models are:
%
\begin{itemize}
\item \textbf{Multi-Layer Perceptron (MLP)}. We trained a four-layer network with two inputs and four outputs.
\item \textbf{ResNet}. Starting with a pretrained ResNet-50 model~\cite{7780459}, the last Bottleneck block in the fourth layer of the network was retrained along with the fully connected layer in an effort to examine if overfitting can be avoided.
\item \textbf{ResNet+LSTM}. An LSTM block was integrated after the average pooling layer of the final residual block of the aforementioned ResNet-50 model in an effort to examine if temporal correlation features of reaching actions can be captured.
\item \textbf{LSTM with Optflow (O-LSTM)}. A single-layer LSTM with 50\% dropout was trained to leverage the information provided by the optical flow images.
\end{itemize}
%
%\item \textbf{3D-CNN}. We created and trained a custom 3D network containing two 3-convolutional blocks, in an effort to examine if the spatio-temporal relations of reaching and no reaching actions between frames can be learned.
%
%\item \textbf{C3D}. We trained a C3D network~\cite{Tran2015LearningSF}, which contains 8 3D convolution layers and 2 fully connected layers followed by a softmax output layer.

All models except for O-LSTM use RGB images of size 224x224 directly as inputs and  were evaluated on our dataset and were trained with learning rate 0.001 using Adam optimizer and cross entropy loss. Both ResNet and ResNet+LSTM model were evaluated with the data augmentation by altering randomly the images through techniques including shift, scale and rotation. 
The O-LSTM uses optical flow images obtained with the Farneback method \cite{farneback2003two} and trained with a learning rate of 0.0001 using Adam optimizer and cross entropy loss and inputs of a flattened image of size 1x12,288 (from a reduced-size 64x64 RGB source image) to reduce the training time.

%%%%%%%%%%%%%%%%%%%%%%%%%%%%%%%%%%%%%%%%

%\sout{MLP and \kkr{RESNET?} use RGB images directly as inputs. The O-LSTM results were obtained using optical flow images as inputs \kkr{by processing RGB images ... need to include these above} }

%\kkm{The aforementioned structures were selected in an effort to strike a balance between network size ...}

%%%%%%%%%%%%%%%%%%%%%%%%%

\subsection{The Proposed BabyNet Algorithm}

\begin{figure}[!h]
\vspace{-10pt}
    \centering
    \includegraphics[scale = 0.55]{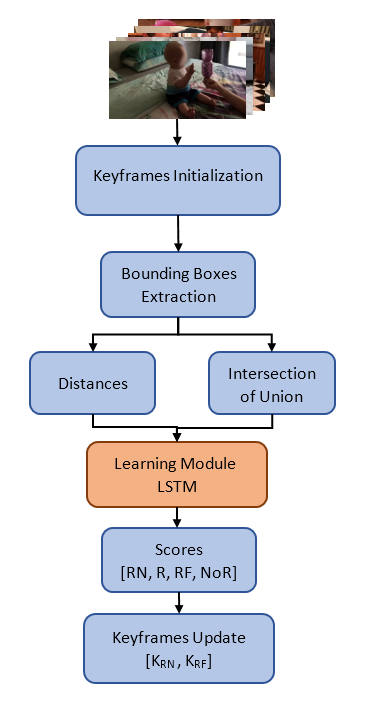}
    \vspace{-6pt}
    \caption{The underlying process followed by our proposed \emph{BabyNet} network for infant reaching action recognition.}
    \label{fig:Algo}
\end{figure}

The flow chart of our proposed algorithm for infant reaching action recognition (\emph{BabyNet}) is depicted in Figure~\ref{fig:Algo}. 
%
%The RN and RF annotations were utilized at separate stages for validation of the proposed algorithm.
%
Given sequence of $\mathsf{T}$ video frames, let $\mathsf{X} = \{x_i^{\mathsf{T}}\}$ and $\mathsf{H} = \{h_L^{\mathsf{T}},h_R^{\mathsf{T}}\}$ denote the objects and left and right hands detected in the frame sequence, respectively. 
Note that we did not incorporate any tracking information, but instead we relied on the centers of the bounding boxes as the main source of information regarding the hands' and objects' locations on the image. 
The coordinates of the bounding boxes were acquired through object detection along with the labels for each frame $\mathsf{T}_{i}$. The two keyframes, $\mathsf{T}_{RN}$ for the onset and $\mathsf{T}_{RF}$ for the offset, are initialized at the first frame.
To enable spatial-temporal reasoning between $\mathsf{X}$ and $\mathsf{H}$, we elected to split a reaching action into two distinct phases: onset and offset. Each phase was detected through a separate process.

%%%%%%%%%%%%%%%%%%%%%%%%%%%%%%%%%%%%%%%
%%%%%%%%%%%%%%%%%%%%%%%%%%%%%%%%%%%%%%%%

%our aim is to perform spatial-temporal reasoning between $\mathsf{X}$ and $\mathsf{H}$ by dividing the action into the two stages: onset and offset as each will be detected as follows:

\begin{table*}[!h]
\vspace{6pt}
\caption{Comparative Results of the Performance the Network Structures Considered in this Work.} 
\vspace{-10pt}
\begin{center}
{\small
\begin{tabular}{p{2.25cm}>{\centering}p{3.5cm}>{\centering}p{2.0cm}>{\centering}p{2.0cm}>{\centering}p{2.0cm}>{\centering}p{1.5cm}>{\centering\arraybackslash}p{1.5cm}}
\toprule
\multirow{2}{*}{Model} & Parameters & Avg. Training & Avg. Validation & Avg. Testing & Precision & Recall \\ 
& (Trainable)$/$(Total) & Accuracy [\%] & Accuracy [\%] & Accuracy [\%] & NoR / R & NoR / R \\
\midrule
\midrule
ResNet %Balanced Dataset No DA 
& \multirow{2}{*}{4,468,739~/~23,514,179} & \multirow{2}{*}{98.30} & \multirow{2}{*}{50.61} & \multirow{2}{*}{53.43} &
\multirow{2}{*}{0.65~/~0.40} &
\multirow{2}{*}{0.55~/~0.51} \\ 
No Data Augm. & & & & & &\\
\midrule
ResNet & \multirow{2}{*}{4,468,739~/~23,514,179} & \multirow{2}{*}{94.59} & \multirow{2}{*}{53.65} & \multirow{2}{*}{58.16} &
\multirow{2}{*}{0.68~/~0.45} &
\multirow{2}{*}{0.62~/~0.52} \\ 
With Data Augm. & & & & & &\\ 
\midrule
%
%ResNet, Imbalanced Dataset & \multirow{2}{*}{4,468,739 ~ / ~23,514,179} & \multirow{2}{*}{97.69} & \multirow{2}{*}{99.06} & \multirow{2}{*}{99.82} \\ 
%With Data Augmentation & & & &\\ 
%
%\midrule
%
ResNet+LSTM & \multirow{2}{*}{9,186,819~/~28,232,259} & \multirow{2}{*}{98.61} & \multirow{2}{*}{50.59} & \multirow{2}{*}{54.21} &
\multirow{2}{*}{0.65~/~0.41} &
\multirow{2}{*}{0.57~/~0.49} \\ 
No Data Augm. & & & & & &\\
\midrule
ResNet+LSTM & \multirow{2}{*}{9,186,819~/~28,232,259} & \multirow{2}{*}{94.31} & \multirow{2}{*}{54.53} & \multirow{2}{*}{54.42} &
\multirow{2}{*}{0.68~/~0.42} &
\multirow{2}{*}{0.51~/~0.60} \\ 
With Data Augm. & & & & & &\\
%Imbalanced Dataset & & & \\
%
\midrule
%
%\multirow{2}{*}{3D-CNN} & \multirow{2}{*}{202,394,946 ~ / ~202,394,946} & %\multirow{2}{*}{89.64} & \multirow{2}{*}{58.33} & \multirow{2}{*}{80} \\
%& & & \\
%
%\midrule
%
%\multirow{2}{*}{C3D} & \multirow{2}{*}{222,597,351 ~ / ~ 222,597,351} & \multirow{2}{*}{55.85} & \multirow{2}{*}{70} & \multirow{2}{*}{70} \\ 
%& & & \\
%
%\midrule
%
%
\multirow{2}{*}{O-LSTM} & \multirow{2}{*}{117,460,994 ~/~ 117,460,994} & \multirow{2}{*}{75.44} & \multirow{2}{*}{81.16} & \multirow{2}{*}{63.71} &
\multirow{2}{*}{0.59~/~0.82} &
\multirow{2}{*}{0.92~/~0.35} \\ 
& & & & & &\\
\midrule
\multirow{2}{*}{MLP} & \multirow{2}{*}{144~/~144} & \multirow{2}{*}{47.66} & 
\multirow{2}{*}{46.13} &
\multirow{2}{*}{51.8} &
\multirow{2}{*}{0.55~/~0.67} &
\multirow{2}{*}{0.78~/~0.42} \\ 
& & & & & &\\
\midrule
\multirow{2}{*}{\emph{BabyNet} (Ours)} & \multirow{2}{*}{1,204~/~1,204} & 
\multirow{2}{*}{44.45} &
\multirow{2}{*}{38.93} &
\multirow{2}{*}{66.27} &
\multirow{2}{*}{0.57~/~0.66} &
\multirow{2}{*}{0.72~/~0.49} \\ 
& & & & & &\\
\bottomrule
\end{tabular}
}
\end{center}\label{tab:Table2}
\vspace{-18pt}
\end{table*}
%%%%%%%%%%%%%%%%%%%%%%%%%%%%%%%%%%%%%%%%%%%%%%%%%%%%%%%%%%%%%%%%%%%%

\subsubsection{The Reaching Onset Phase}
Using the coordinates of the bounding boxes, we can first compute the distance $d^i_j$ between the hands $\{h_L^{\mathsf{T}},h_R^{\mathsf{T}}\}$ and the object $x_i$ in  $\mathsf{X}$ in the current frame $\mathsf{T}_j$. %We compare $d^i_j$ with $d^i_{j-1}$, the distance obtained from $\mathsf{T}_{j-1}$. 
When $d^i_{j}-d^i_{j-1} < 0$, the onset detection remains valid and hence $\mathsf{T}_{RN}$ can be kept as a keyframe. When $d^i_{j}-d^i_{j-1} \geq 0$ and that positive increase continues for four consecutive frames, the onset is invalidated and $\mathsf{T}_{i}$ can be set as the new keyframe $\mathsf{T}_{RN}$. This approach can help avoid false negatives during this phase.

\subsubsection{The Reaching Offset Phase}
The intersection of union $(IOU)$ between the hands $\{h_L^{\mathsf{T}},h_R^{\mathsf{T}}\}$ and the object $x_i$ can be estimated and compared to a threshold value determined through the learning process. If the $IOU$ is less than the threshold, no offset will be detected and the keyframe $\mathsf{T}_{RF}$ can be definitively set as $\mathsf{T}_{i}$. Otherwise, the keyframe is updated as $\mathsf{T}_{i}$ until the next frame. 

\subsubsection{Core Structure of BabyNet}
Our proposed \emph{BabyNet} uses the LSTM structure to learn the relation between the bounding boxes through the input consisting of the distances and intersection of union $(IOU)$. The output is the scores for onset (RN), offset (RF), reach (R) and no reach (NoR) are used to update identified keyframes. The reach (R) is the label used for all frames between the onset RN and offset RF frames. Similarly, the no reach (NoR) is the label for all frames before the onset and after the offset. 

\subsubsection{Implementation Details}
\emph{BabyNet} has two inputs $(T=2)$ and four outputs (RN, RF, NoR, R). Preliminary testing showed that selecting $T=2$ can improve temporal correlation predictions without overfitting. \emph{BabyNet} was trained with learning rate 0.001 using Adam optimizer and cross entropy loss. In initial testing we used a small dataset of 63 reaches with a 60\% training, 15\% validation, and 25\% testing split. While \emph{BabyNet} (and MLP) can perform well with small datasets (as desired), larger structures (ResNet variants and O-LSTM) were found to overfit the dataset. To resolve this issue, we tested larger networks with the full dataset (193 reaches) while keeping \emph{BabyNet} and MLP at 63 reaches.%; results reported in the following section

% ---------------------------------- %
\section{Results}\label{results}
% ---------------------------------- %

%Table~\ref{tab:Table2} contains results obtained by the 2D network implementations, MLP and \emph{BabyNet} that uses RGB images as inputs. The O-LSTM results were obtained using optical flow images as inputs.  % that use still images/video frames as input along with the results obtained by evaluating the 3D network implementations (including our proposed \emph{BabyNet} structure) that use sequences of video frames as input.
%%In contrast,Table~\ref{tab:Table4} contains the results obtained by evaluating the 3D network implementations (including our proposed \emph{BabyNet} structure) that use sequences of video frames as input.

Comparison results, including classification accuracy (range: $[0-100 \%]$) and precision/recall scores (range: $[0-1]$), are shown in Table~\ref{tab:Table2}. Including the precision/recall scores of both no reaches (NoR) and reaches (R) serves a dual purpose: (1) to examine the trade-off between different methods; and (2) to help attain a more clear distinction in terms of the structures' capability to correctly differentiate between no reaches and reaches.

We first compared the performance of the family of ResNet and ResNet+LSTM structures. The ResNet structure with data augmentation achieves the best performance with an average testing accuracy of $58.16 \%$.
With reference to Table~\ref{tab:Table2}, we can observe that the models with data augmentation give nearly the same results as with those without data augmentation (cf. $58.16 \%$ to $53.43 \%$ in ResNet and $54.42 \%$ to $54.21 \%$ in ResNet+LSTM). 
To properly gauge the effect of data augmentation on accuracy, we compared the trade-off between precision and recall of the two networks. The ResNet model with data augmentation has higher precision along with the ResNet+LSTM model with data augmentation. However, the recall score is slightly different as the ResNet structure with data augmentation achieved $0.62$ for no reaches (NoR) and $0.52$ for reaches (R) which indicates that the number of false negatives is lower for the no reaches (NoR) cases. In contrast, the ResNet+LSTM with data augmentation had a recall of $0.51$ for no reaches and $0.60$ for reaches, thus leading to a lower number of false negatives for the case of reaches (Table~\ref{tab:Table2}).

%\sout{However, results indicate that despite the high accuracy obtained for the 2D-ResNet, 2D network structures have a very low precision, and reach average accuracy when tested with a sequence of video frames as input (Table~\ref{tab:Table3}). They can differentiate between reaching and not reaching instances, but cannot switch from one to another correctly when a complete video (i.e. not trimmed to clearly contain when the Reaching Onset and Reaching Offset phases occur) is presented to the networks. %The ablation study in the next section discusses the effect of critical parameters such as
%showing that the network fails to the task . 

%For the 3D networks, in order to utilize our dataset for training, the first step is to ensure the classes have the same number of frames for the duration of the action. We modified the individual frame rate or frames per second (fps) for each of the videos and upsampled all videos at a fixed value of frame. To find the optimal value of upsampling, we analyzed the frequency and the duration of the reaches and identified the maximum number of frames (63 frames) in order to avoid loss of frames (Figure~\ref{fig:histogram2}).}

\begin{figure}[!h]
\vspace{6pt}
    \centering
    \includegraphics[scale = 0.39]{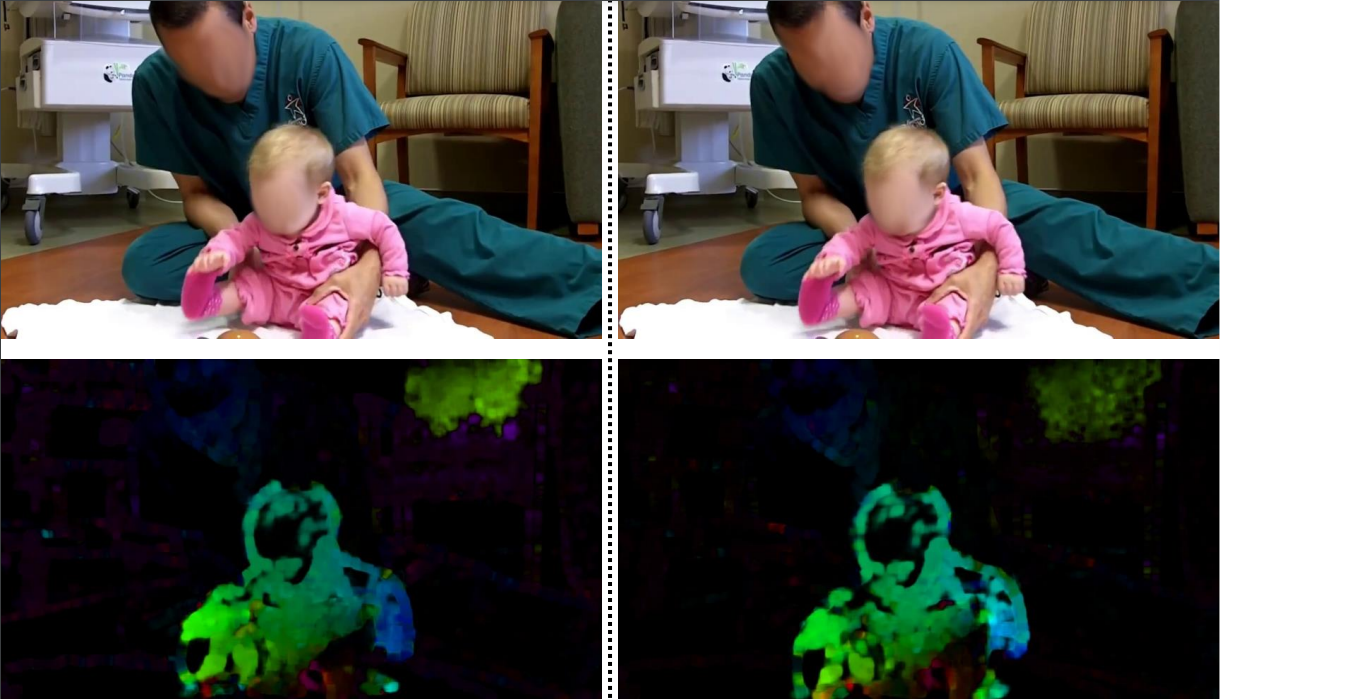}
    \vspace{-20pt}
    \caption{RGB (top panels) and equivalent optical flow images (bottom panels) spaced two frames apart (from left to right). Optical flow images can capture more clearly small and subtle changes, but at a (significant) pre-processing computational cost compared to plain RGB image inputs.}
    \label{fig:olstm}
    \vspace{-10pt}
\end{figure}

Results of the O-LSTM trained with optical flow images show that this structure can achieve $63.71\%$ of average testing accuracy, surpassing all ResNet-based structures. In terms of precision and recall, the model has lower false positive detection for reaches (R) detection than no reaches (NoR) ones (cf. $0.59$ to $0.82$), but these detections are more likely to be false negatives in the case of reaches (given the recall score of $0.35$)---it was a challenge to learn to recognize short reaches. %as it can't detect a range of small motion. . 
O-LSTM requires a significantly larger number of parameters since the input optical flow image needs to be flattened (i.e. transform from a size of 3x64x64 to 1x12,288). This is in addition to the necessary pre-processing (high) computational effort required to transform the original RGB image to optical flow image (see Figure~\ref{fig:olstm}).

Results demonstrate that our proposed \emph{BabyNet} outperforms all structures in terms of average testing accuracy while using the second smallest number of parameters and the small dataset of 63 reaches and in spite of featuring lower average training/validation accuracy. The observed lower validation accuracy can be explained by the fact that more challenging reaches were included in the validation set which, nevertheless, did not impact the ability of \emph{BabyNet} 
to learn pertinent features. 
The observed lower training accuracy can be associated with the reduced training dataset size, whereby training accuracy would keep improving with more data. %In contrast to larger ResNet network variants (which are overfitting),
Still, \emph{BabyNet} can perform well (in terms of testing accuracy) despite sub-optimal training.

%%%%%%%%%%%%%%%%%%%%%%%%%%%%%%%%%%%%%%%

The MLP structure is the smallest one (also uses 63 reaches), but it has the worst average testing accuracy (see Table~\ref{tab:Table2}). %based on the testing accuracy for the same split ($66.27 \%$ vs $51.8\%$); see Table~\ref{tab:Table2}. 
%In fact, \emph{BabyNet} is able to learn more pertinent features regardless of its low accuracy results. 
Furthermore, both networks predicted the same number of frames incorrectly during the reach phase but the MLP predicted 20 frames incorrectly during the no reach action whereas \emph{BabyNet} only predicted six frames incorrectly. In term of keyframes, \emph{BabyNet} only had a delay of one frame while predicting the reach whilst the MLP had a delay of four frames. 
In contrast, the MLP was not able to learn the motion of the reaching action, and had difficulties to discern the transition from a reach to a no reach.

\section{Discussion and Conclusions}
% ---------------------------------- %
%\paragraph*{\textbf{Contributions and Key Findings}} 
The paper proposed a new light-weight network called \emph{BabyNet} aimed at infant reaching action recognition. Our approach was found able to model short-range and long-range motion correlation of different key phases of a reaching action: its onset and offset. 
To develop the network and evaluate its performance, we also developed a dataset specifically suitable for infant action recognition that includes action and bounding box annotations. %Datasets like this one are limited in the literature.

Evaluation results demonstrated that our proposed \emph{BabyNet} is small yet powerful, and can challenge the performance of significantly larger structures by achieving $66.27\%$ average testing accuracy (the highest one) on our dataset. The family of Resnet-based structures, despite their solid performance during training and validation, were found to provide results with increased false positives rates. On the other hand, the O-LSTM (that has the second best average testing accuracy and comparable to \emph{BabyNet}'s) could not balance between no reaches (NoR) and reaches (R)---recall rates of 0.92 and 0.35, respectively. Yet, it remains an approach worth of further investigation in future work due to the ability of optical flow images to better differentiate subtle motion patterns compared to RGB images (see Figure~\ref{fig:olstm}). %On the other hand, the 3D networks provided sustainable results being  tested on cropped videos including reaching and no reaching action but were unable to perform with untrimmed videos.

Compared to the MLP (which is of comparable size to \emph{BabyNet}), our \emph{BabyNet} performed much better (approximately $27$\% improved performance) with almost the same precision/recall scores. However, it provides onset and offset keyframes at precision of one frame while the MLP had a delay of 4 frames. 
These promising findings demonstrate that \emph{BabyNet} can serve as a light-weight data-driven framework for video-based infant reaching action recognition.

%\paragraph*{\textbf{Directions for Future Work}}
As this stage, \emph{BabyNet} can be challenged by the lack of viewpoint variation in the dataset and reliance on the performance of the detector network; missing detection of the hands and/or the object can undermine reaching action recognition. 
Future work aims to extend the dataset introduced herein to help generalize to reaching action in other postures. 
As mentioned above, these efforts gear toward the implementation and assessment of the \emph{BabyNet}'s performance in closing the action-perception loop in our upper extremity pediatric wearable robotic device under development~\cite{Kokkoni2020DevelopmentReachingb}.

\bibliographystyle{IEEEtran}
\bibliography{AmelREFS,ElenaREFS_NoMendeley}

\end{document}